# COVID-19-related Nepali Tweets Classification in a Low Resource Setting

**Rabin Adhikari**[1,2], **Safal Thapaliya**[1,2], **Nirajan Basnet**[1,2], **Samip Poudel**[1,2]
**Aman Shakya**[2], **Bishesh Khanal**[1]

[1]NepAl Applied Mathematics and Informatics Institute for research (NAAMII)
[2]Institute of Engineering, Pulchowk Campus, Tribhuvan University

## Abstract

Billions of people across the globe have been using social media platforms in their local languages to voice their opinions about the various topics related to the COVID-19 pandemic. Several organizations, including the World Health Organization, have developed automated social media analysis tools that classify COVID-19-related tweets into various topics. However, these tools that help combat the pandemic are limited to very few languages, making several countries unable to take their benefit. While multi-lingual or low-resource language-specific tools are being developed, they still need to expand their coverage, such as for the Nepali language. In this paper, we identify the eight most common COVID-19 discussion topics among the Twitter community using the Nepali language, set up an online platform to automatically gather Nepali tweets containing the COVID-19-related keywords, classify the tweets into the eight topics, and visualize the results across the period in a web-based dashboard. We compare the performance of two state-of-the-art multi-lingual language models for Nepali tweet classification, one generic (mBERT) and the other Nepali language family-specific model (MuRIL). Our results show that the models' relative performance depends on the data size, with MuRIL doing better for a larger dataset. The annotated data, models, and the web-based dashboard are open-sourced at `https://github.com/naamiinepal/covid-tweet-classification`.

## 1 Introduction

The COVID-19 pandemic has caused a global rise in social media users who express their opinions and share information on various topics related to the pandemic. Public health organizations and relevant agencies could analyze the social media data for early warning on potentially new virus variants based on symptoms discussion, for understanding the impact of various intervention measures, the efficacy of vaccination programs, etc. Social media data analysis can help develop strategies for combating the pandemic (Yigitcanlar et al., 2020), and improve the efficiency of the health industry (Scanfeld et al., 2010; Signorini et al., 2011; Harris et al., 2013; Paul and Dredze, 2014; Eichstaedt et al., 2015).

Several studies performed sentiment analysis of tweets to understand people's views towards the pandemic (Dubey, 2020; Jelodar et al., 2020; Samuel et al., 2020; Alamoodi et al., 2021). Since sentiment analysis provides limited coarse-level information, recently, there has been an interest in building tools for early warning and topic-level discourse analysis. Most notably, the World Health Organization (WHO) tracks internet discourse by examining global pandemic-related Twitter data and news using tools like COVID-19 News Map[1] and EARS[2]. Although a significant fraction of the global population uses local languages in social media, most of these tools are limited to English or Anglo-European languages. For instance, the WHO EARS works in only nine languages, piloted in 30 countries.

In recent years, there has been a growing interest in building multi-lingual language models, building low-resource language datasets, and exploring NLP methods with smaller language models and smaller data (Conneau et al., 2019; Wang et al., 2020; Ogueji et al., 2021). Nepali is a low-resource language with a significant gap in advances, data availability, and the development of NLP tools. While there has been some work on low-resource languages for sentiment analysis in low-resource languages (Addawood et al., 2020; Hosseini et al., 2020) including Nepali (Sitaula et al., 2021; Shahi et al., 2022), to our knowledge there is no work on COVID-19 tweet topics classification for discourse

---

[1] `https://portal.who.int/eios-coronavirus-newsmap/`
[2] `https://www.who-ears.com/`

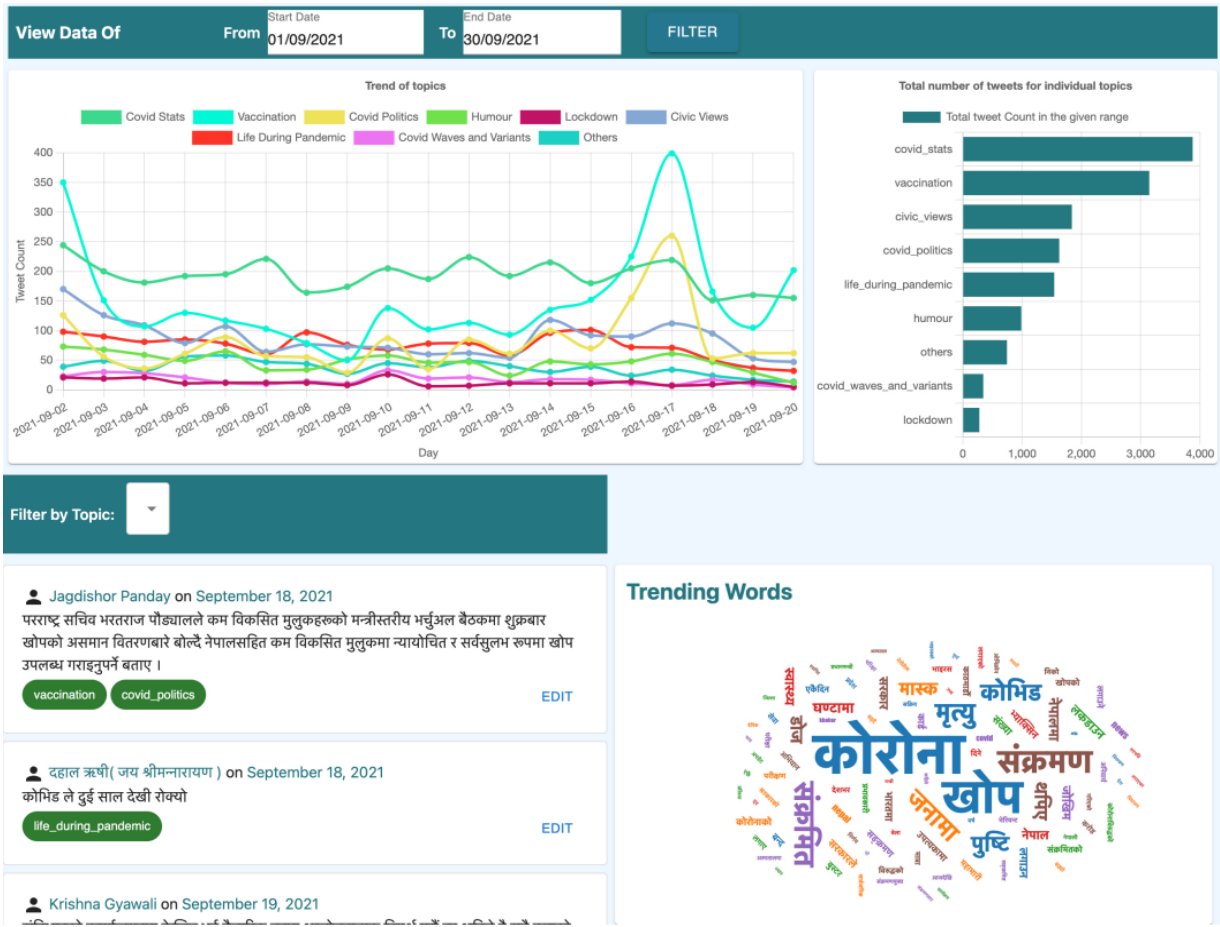

Figure 1: The web app dashboard shown above uses infographics, viz. bar graphs, and line charts, to track the trend of various topics. We can filter the tweets based on time and topics to make analysis easier. Additionally, we have incorporated humans in the loop by developing an administrator interface to validate the predicted tweet labels from the model and proofread the validated ones.

analysis in the Nepali language.

In this work, we propose a new dataset, deep learning classification models based on multilingual language models, and an interactive dashboard for incremental learning and visualization of COVID-19 tweets topic classification in the Nepali language. Figure 1 shows a snapshot of our dashboard. In addition to visualizing topics classification in real-time, the dashboard can manually verify the ML model's prediction, correct the predictions to annotate more data, and retrain the model via GUI for improvement as more data becomes available.

The followings are our contributions to the scientific community.

- We release a multi-annotator multi-label Nepali Annotated Tweets with COVID-19 Topics Classification (NAT-CTC) dataset that contains 12, 241 tweets in Devanagari script, manually tagged with eight simplified topics. We also provide inter-annotator agreement results on this dataset using four annotators labeling 400 typical tweets.

- We release our open-source web-based platform with GUI for automatic keywords-based tweets collection, tweet pre-processing, topic classification, and visualization. This platform can be used for AI-assisted annotation and incremental learning, where human annotators can correct the labels predicted by ML models and then retrain ML models. We use this approach during the dataset preparation as well.

- We show that the benefit of using a Nepali language family-specific model compared to generic multi-lingual language models may come only if there is a certain minimum number of annotated data for the downstream task.

- We analyze 98, 849 tweets using our topic classification model and the dashboard and show how the frequency of discussion in eight topics varied over time during the pandemic.

## 2 NAT-CTC Dataset

### 2.1 Keyword-based Filter for Tweet Collection

We identified 48 keywords in the Devanagari script that cover the majority of COVID-19-related tweets and used *twarc*[3] to extract tweets that contain the keywords and is tagged as Nepali by Twitter. New keywords were iteratively added into an initial set using *twarc* and manual review until finally settling on 48 keywords.

### 2.2 Eight COVID-19-related Topics

While the WHO EARS has 30 topics, to reduce the complexity and due to very limited tweets for Nepali language in some topics, we contextualized and developed these eight topics suitable to describe the specific narratives in Nepal: *COVID Stats*, *Vaccination*, *COVID Politics*, *Humor*, *Lockdown*, *Civic Views*, *Life during Pandemic*, and *Waves and Variants*.

### 2.3 Multi-annotator Manual Annotation with Incremental Learning

Seven annotators initially used *Label Studio*[4] to annotate the tweets (one tweet annotated by only one person), after which we trained a machine learning model (see subsection 3.2) to predict labels that were then corrected by the annotators using our dashboard. This increased annotation speed substantially and improved the ML model as more data came in. Each tweet could be tagged with multiple topics. Moreover, we randomly selected 400 tweets to study inter-rater agreement, each of which was annotated by four annotators. The tweets chosen for the agreement are randomly sampled from the labeled dataset. Their exact number and the proportion to the larger labeled dataset can be seen from Figure 2.

## 3 Methods

### 3.1 Tweets Pre-processing

The tweets were pre-processed in the given order using *Pandas*[5]:

---
[3]https://github.com/DocNow/twarc
[4]https://labelstud.io/
[5]https://pandas.pydata.org/

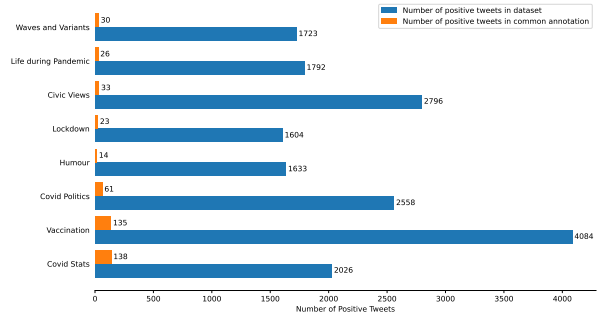

Figure 2: number of positive tweets for each label. The imbalanced nature of the multi-label classification can be seen. *Vaccination* has only 4084 positive samples out of 12, 241 tweets; other topics are even more imbalanced.

1. Remove user mentions and links and change Latin characters to lowercase.

2. Reduce spaces to a single space between words.

3. To eliminate bias from source information, remove text followed by the word *via*.

4. Remove leading and trailing spaces and tweets with three or fewer words.

5. Normalize the Unicode strings to NFKC standards[6].

### 3.2 Tweet's Topic Classification

We utilized Indic Language multi-lingual model MuRIL's (Khanuja et al., 2021) preprocessing and encoder models with a batch normalization (Ioffe and Szegedy, 2015) layer and a linear classifier with a dropout rate of 0.5 to categorize the preprocessed tweets into the eight topics. For the training approach, we followed the blog, *A Recipe for Training Neural Networks*[7]. We adjusted the initial bias of the output layer to $\log\left(\frac{pos}{neg}\right)$ to reflect the imbalance in the dataset and facilitate the initial convergence of the model. We utilized the AdamW (Loshchilov and Hutter, 2017) with 0.01 weight decay, a learning rate of $5 \times 10^{-5}$ and used the Polynomial Decay Scheduler with 10% of the total training steps as Warmup[8]. Since Precision and Recall are unaffected by class imbalance (Saito and Rehmsmeier, 2015; Branco et al., 2016), the preferred metrics to evaluate the model's performance

---
[6]https://unicode.org/reports/tr15/
[7]https://karpathy.github.io/2019/04/25/recipe/
[8]https://github.com/tensorflow/models/blob/v2.7.2/official/nlp/optimization.py

| Labels | F1 Score | Area under PR Curve | Fleiss' Kappa Score |
|---|---|---|---|
| COVID Stats | 0.913 ± 0.008 | 0.964 ± 0.003 | 0.87 |
| Vaccination | 0.974 ± 0.002 | 0.984 ± 0.002 | 0.88 |
| COVID Politics | 0.711 ± 0.012 | 0.763 ± 0.013 | 0.42 |
| Humor | 0.737 ± 0.014 | 0.766 ± 0.016 | 0.65 |
| Lockdown | 0.967 ± 0.005 | 0.988 ± 0.004 | 0.79 |
| Civic Views | 0.729 ± 0.007 | 0.757 ± 0.01 | 0.61 |
| Life During Pandemic | 0.61 ± 0.03 | 0.616 ± 0.043 | 0.34 |
| Waves and Variants | 0.851 ± 0.006 | 0.915 ± 0.005 | 0.53 |

Table 1: Area under PR-curve and F1-Score for each label, along with the corresponding Fleiss' Kappa score. In addition to depicting the mean value for each metric and its standard deviation for 5-fold cross-validation, it helps to find the correlation between the metrics and the corresponding Kappa score.

were the F1 score with weighted averaging and the Area under the PR Curve (AUPR).

## 4 Experiments and Results

### 4.1 Topics Inter-rater Agreement

With 400 tweets annotated by four annotators each, we calculated Fleiss' Kappa (Fleiss, 1971; Fleiss et al., 2013) score for each of the eight categories. Since our dataset contains multi-label classification, we have reported the Kappa score for the individual categories (shown in Table 1). We averaged the Fleiss' Kappa scores of the individual categories to obtain a mean of 0.64, which shows substantial agreement between the four annotators (Artstein and Poesio, 2008; McHugh, 2012).

### 4.2 Label-wise Model Performance

Table 1 shows mean and standard deviation of model performance for 5-fold cross-validation where each fold consisted of 2,448 tweets. From the inter-rater agreement Kappa score for the eight labels shown aside, we can infer that the performance seems to drop as the agreement among the annotators for the target class reduces (reduced Kappa score), with a few exceptions such as *Waves and Variants*.

| Avg. Type | F1 Score | AUPR |
|---|---|---|
| Micro | 0.817 ± 0.005 | - |
| Macro | 0.811 ± 0.004 | 0.841 ± 0.007 |
| Weighted | 0.823 ± 0.004 | 0.854 ± 0.006 |

Table 2: Micro, Macro, and Weighted scores to capture the overall performance across all the target topics.

Table 2 presents the overall performance across all the topics using various averaging of AUPR and F1 Score[9]. The model's prediction for some tweets is shown in Table 3.

### 4.3 Performance of mBERT and MuRIL Changes Differently When Training Dataset Size Changes

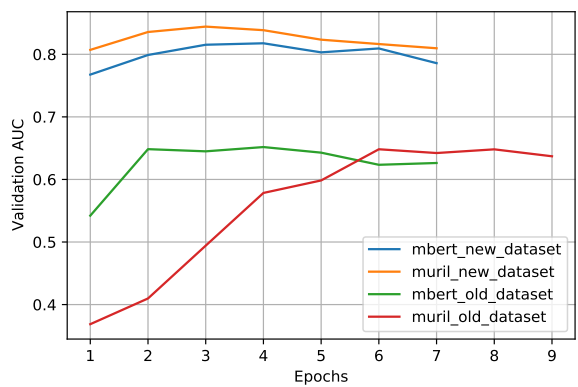

Figure 3: MuRIL performed better than mBERT when dataset size was increased from 6,952 to 12,241. For the smaller dataset, MuRIL has 0.64 mean AUPR, whereas mBERT has 0.65. MuRIL has 0.84 mean AUPR for the larger dataset, whereas mBERT has 0.81.

We compared mBERT (Devlin et al., 2019) and MuRIL (Khanuja et al., 2021) for the same normalization and dropout for various training data sizes. As shown in Figure 3, MuRIL had slightly lower performance on a smaller dataset of 6,952 tweets but outperformed mBERT by a more significant margin when the dataset size increased to 12,241 tweets. The language family-specific models provide a more significant benefit than generic models only when finetuning training dataset size is of a certain minimum number. However, this has to be further explored with additional languages and language models.

---
[9] https://scikit-learn.org/stable/modules/generated/sklearn.metrics.f1_score.html

| Example | English Translation | Label Prediction |
|---|---|---|
| कोरोनाको परीक्षण किन घटाउँदै छ नालायक सरकार? | Why is the worthless government reducing the testing of Corona? | *COVID Politics*, *Civic views* |
| पछिल्लो २४ घण्टामा थप ९४४ जनामा कोरोना संक्रमण पुष्टि, संक्रिय संक्रमितको संख्या ६ हजार नाघ्यो | In the last 24 hours, 944 more Corona infections have been confirmed, and the number of active infected has exceeded 6 thousand | *COVID Stats* |
| कोरोनाको ग्राफ उकालो लागि सक्यो, बेलैमा झार्नु पर्यो, भिडभाडमा नजाने र मास्क लगाउने नै प्रमुख उपाय। | The graph of Corona has gone up, it has to be realized early, the best solution is not to go to crowded places and wear a mask. | *Civic Views*, *Life during Pandemic*, *Waves and Variants* |
| फेरि लकडाउनको हल्लाले उद्योगी चिन्तित, कोरोना खोप लगाएर व्यवसाय सन्चालन गर्न पाउनुपर्ने माग | Again, the rumors of the lockdown are worrying the industrialists, they demand to be able to operate the business with Corona vaccination. | *Vaccination*, *Lockdown*, *Life during Pandemic* |
| तेत्रो लकडाउन त एक्लै कटाइयो जाबो February 14 एक दिन त कसो कटाउन नसकिएला र ? | That lockdown was spent alone, I can spend February 14 alone, can't I? | *Humour*, *Lockdown* |

Table 3: Examples of some model predictions

### 4.4 Trend Analysis

We present here a couple of examples of insights from our ML model's classification results on 98,849 tweets from June 2021 to February 2022. Discussions related to *Vaccination* surged in mid-September 2021, which was just after Nepal's Government decided to provide vaccines to students and youths[10], as can be seen in Figure 1. Similarly, tweets related to *COVID Stats* increased during late January and early February of 2022 when Nepal was hit by the third wave of the virus[11].

## 5 Discussion and Conclusion

We have developed an online platform to gather, analyze, and categorize tweets about COVID-19 in Nepali, written in the Devanagari script. We selected eight topics pertinent to online discussions in Nepal based on the various categories of online discourse established by the WHO. Seven different people have annotated 12,241 tweets from our dataset. We arrived at our best model architecture using MuRIL (Khanuja et al., 2021) as the encoder and a batch normalization (Ioffe and Szegedy, 2015) layer immediately before the final output layer after finetuning several hyperparameters and utilizing two encoder backbones, i.e., mBERT (Devlin et al., 2019) and MuRIL (Khanuja et al., 2021). The web app dashboard uses infographics like bar graphs and area charts to track the development of online discussions. We can filter the tweets based on time and topics to make analysis easier. Additionally, we developed an administrator interface to validate the predicted tweet labels from the model and proofread the validated ones.

Although our dataset is not large, it may be valuable for transfer learning and semi-supervised learning (Lwowski and Najafirad, 2020). Our dataset can help to make multi-lingual datasets more inclusive and the models trained on them more robust. Translating and transliterating to and from our dataset can help in augmentation in various settings.

---
[10] https://kathmandupost.com/health/2021/09/20/all-students-above-18-to-be-jabbed-with-covid-19-vaccine
[11] https://english.onlinekhabar.com/nepal-covid-19-third-wave-signs.html

# References

Aseel Addawood, Alhanouf Alsuwailem, Ali Alohali, Dalal Alajaji, Mashail Alturki, Jaida Alsuhaibani, and Fawziah Aljabli. 2020. Tracking and understanding public reaction during COVID-19: Saudi Arabia as a use case. In *Proceedings of the 1st Workshop on NLP for COVID-19 (Part 2) at EMNLP 2020*, Online. Association for Computational Linguistics.

Abdullah Hussein Alamoodi, Bilal Bahaa Zaidan, Aws Alaa Zaidan, Osamah Shihab Albahri, KI Mohammed, Rami Qays Malik, Esam Motashar Almahdi, Mohammed A Chyad, Ziadoon Tareq, Ahmed Shihab Albahri, et al. 2021. Sentiment analysis and its applications in fighting covid-19 and infectious diseases: A systematic review. *Expert systems with applications*, 167:114155.

Ron Artstein and Massimo Poesio. 2008. Inter-coder agreement for computational linguistics. *Computational linguistics*, 34(4):555–596.

Paula Branco, Luís Torgo, and Rita P Ribeiro. 2016. A survey of predictive modeling on imbalanced domains. *ACM Computing Surveys (CSUR)*, 49(2):1–50.

Alexis Conneau, Kartikay Khandelwal, Naman Goyal, Vishrav Chaudhary, Guillaume Wenzek, Francisco Guzmán, Edouard Grave, Myle Ott, Luke Zettlemoyer, and Veselin Stoyanov. 2019. Unsupervised cross-lingual representation learning at scale. *arXiv preprint arXiv:1911.02116*.

Jacob Devlin, Ming-Wei Chang, Kenton Lee, and Kristina Toutanova. 2019. BERT: Pre-training of deep bidirectional transformers for language understanding. In *Proceedings of the 2019 Conference of the North American Chapter of the Association for Computational Linguistics: Human Language Technologies, Volume 1 (Long and Short Papers)*, pages 4171–4186, Minneapolis, Minnesota. Association for Computational Linguistics.

Akash Dutt Dubey. 2020. Twitter sentiment analysis during covid-19 outbreak. *Available at SSRN 3572023*.

Johannes C Eichstaedt, Hansen Andrew Schwartz, Margaret L Kern, Gregory Park, Darwin R Labarthe, Raina M Merchant, Sneha Jha, Megha Agrawal, Lukasz A Dziurzynski, Maarten Sap, et al. 2015. Psychological language on twitter predicts county-level heart disease mortality. *Psychological science*, 26(2):159–169.

Joseph L Fleiss. 1971. Measuring nominal scale agreement among many raters. *Psychological bulletin*, 76(5):378.

Joseph L Fleiss, Bruce Levin, and Myunghee Cho Paik. 2013. *Statistical methods for rates and proportions*. john wiley & sons.

Jenine K Harris, Nancy L Mueller, and Doneisha Snider. 2013. Social media adoption in local health departments nationwide. *American journal of public health*, 103(9):1700–1707.

Pedram Hosseini, Poorya Hosseini, and David Broniatowski. 2020. Content analysis of Persian/Farsi tweets during COVID-19 pandemic in Iran using NLP. In *Proceedings of the 1st Workshop on NLP for COVID-19 (Part 2) at EMNLP 2020*, Online. Association for Computational Linguistics.

Sergey Ioffe and Christian Szegedy. 2015. Batch normalization: Accelerating deep network training by reducing internal covariate shift. In *International conference on machine learning*, pages 448–456. PMLR.

Hamed Jelodar, Yongli Wang, Rita Orji, and Shucheng Huang. 2020. Deep sentiment classification and topic discovery on novel coronavirus or covid-19 online discussions: Nlp using lstm recurrent neural network approach. *IEEE Journal of Biomedical and Health Informatics*, 24(10):2733–2742.

Simran Khanuja, Diksha Bansal, Sarvesh Mehtani, Savya Khosla, Atreyee Dey, Balaji Gopalan, Dilip Kumar Margam, Pooja Aggarwal, Rajiv Teja Nagipogu, Shachi Dave, et al. 2021. Muril: Multilingual representations for indian languages. *arXiv preprint arXiv:2103.10730*.

Ilya Loshchilov and Frank Hutter. 2017. Decoupled weight decay regularization. *arXiv preprint arXiv:1711.05101*.

Brandon Lwowski and Peyman Najafirad. 2020. COVID-19 surveillance through Twitter using self-supervised and few shot learning. In *Proceedings of the 1st Workshop on NLP for COVID-19 (Part 2) at EMNLP 2020*, Online. Association for Computational Linguistics.

Mary L McHugh. 2012. Interrater reliability: the kappa statistic. *Biochemia medica*, 22(3):276–282.

Kelechi Ogueji, Yuxin Zhu, and Jimmy Lin. 2021. Small data? no problem! exploring the viability of pretrained multilingual language models for low-resourced languages. In *Proceedings of the 1st Workshop on Multilingual Representation Learning*, pages 116–126.

Michael J Paul and Mark Dredze. 2014. Discovering health topics in social media using topic models. *PloS one*, 9(8):e103408.

Takaya Saito and Marc Rehmsmeier. 2015. The precision-recall plot is more informative than the roc plot when evaluating binary classifiers on imbalanced datasets. *PloS one*, 10(3):e0118432.

Jim Samuel, GG Ali, Md Rahman, Ek Esawi, Yana Samuel, et al. 2020. Covid-19 public sentiment insights and machine learning for tweets classification. *Information*, 11(6):314.

Daniel Scanfeld, Vanessa Scanfeld, and Elaine L Larson. 2010. Dissemination of health information through social networks: Twitter and antibiotics. *American journal of infection control*, 38(3):182–188.

TB Shahi, C Sitaula, and N Paudel. 2022. A hybrid feature extraction method for nepali covid-19-related tweets classification. *Computational Intelligence and Neuroscience*, 2022.

Alessio Signorini, Alberto Maria Segre, and Philip M Polgreen. 2011. The use of twitter to track levels of disease activity and public concern in the us during the influenza a h1n1 pandemic. *PloS one*, 6(5):e19467.

Chiranjibi Sitaula, Anish Basnet, A Mainali, and Tej Bahadur Shahi. 2021. Deep learning-based methods for sentiment analysis on nepali covid-19-related tweets. *Computational Intelligence and Neuroscience*, 2021.

Zihan Wang, Stephen Mayhew, Dan Roth, et al. 2020. Extending multilingual bert to low-resource languages. *arXiv preprint arXiv:2004.13640*.

Tan Yigitcanlar, Nayomi Kankanamge, Alexander Preston, Palvinderjit Singh Gill, Maqsood Rezayee, Mahsan Ostadnia, Bo Xia, and Giuseppe Ioppolo. 2020. How can social media analytics assist authorities in pandemic-related policy decisions? insights from australian states and territories. *Health Information Science and Systems*, 8(1):1–21.